\documentclass[12pt,a4paper]{article}
\usepackage[british]{babel}
\usepackage[T1]{fontenc}
\usepackage{tikz, pgfplots}
\usetikzlibrary{positioning, calc, shapes.geometric, arrows, arrows.meta}
\usepackage{cite}
\usepackage[a4paper,top=2cm,bottom=2cm,left=2.25cm,right=2.25cm,marginparwidth=1.75cm]{geometry}
\usepackage{siunitx}
\usepackage{amsmath}
\usepackage{graphicx}
\usepackage[title]{appendix}
\usepackage{mathrsfs}
\usepackage{amsfonts}
\usepackage{booktabs} 
\usepackage{threeparttable} 
\usepackage{algorithm}
\usepackage{algorithmicx}
\usepackage{algpseudocode}
\usepackage{listings}
\usepackage{enumitem}
\usepackage{chngcntr}
\usepackage{lipsum}
\usepackage{subcaption}
\usepackage{authblk}
\usepackage{csquotes}       
\usepackage{diagbox}
\usepackage{caption}
\usepackage{setspace}
\usepackage{titlesec}
\usepackage{lineno}
\usepackage{float}

\usepackage[colorlinks=true, allcolors=blue]{hyperref}

\titleformat{\section}
  {\normalfont\Large\bfseries}{\thesection.}{1em}{}

\rightlinenumbers
\linenumbers


\modulolinenumbers[5]

\title{IO Transformer: Evaluating SwinV2-Based Reward Models for Computer Vision}
\author{Maxwell Meyer}
\author{Jack Spruyt}
\affil{\{maxwellmeyer, jackspruyt\}@pramadevelopment.com}
\date{}  
\pgfplotsset{compat=1.18}

\begin{document}
\nolinenumbers
\maketitle

\begin{abstract}
Transformers and their derivatives have achieved state-of-the-art performance across text, vision, and speech recognition tasks. However, minimal effort has been made to train transformers capable of evaluating the output quality of other models. This paper examines SwinV2-based reward models, called the Input-Output Transformer (IO Transformer) and the Output Transformer. These reward models can be leveraged for tasks such as inference quality evaluation, data categorization, and policy optimization. Our experiments demonstrate highly accurate model output quality assessment across domains where the output is entirely dependent on the input, with the IO Transformer achieving perfect evaluation accuracy on the Change Dataset 25 (CD25). We also explore modified Swin V2 architectures. Ultimately Swin V2 remains on top with a score of \SI{95.41}{\percent} on the IO Segmentation Dataset, outperforming the IO Transformer in scenarios where the output is not entirely dependent on the input. Our work expands the application of transformer architectures to reward modeling in computer vision and provides critical insights into optimizing these models for various tasks.
\end{abstract}

\section{Introduction}
Transformers have emerged as a dominant architecture in numerous domains, including natural language processing (NLP), computer vision, and speech recognition, largely due to their powerful attention mechanisms and ability to accurately model long-range dependencies. Initially introduced for NLP tasks by Vaswani et al. \cite{vaswani2017attention}, transformers have since been adapted to vision with models like the Vision Transformer (ViT) \cite{dosovitskiy2020image} and Swin Transformer \cite{liu2021swin}, which have achieved state-of-the-art performance in tasks such as image classification \cite{touvron2021training}, segmentation \cite{hatamizadeh2021swin}, and object detection \cite{liu2021swin}. Despite these advances, there has been limited exploration into using transformer architectures for evaluating the quality of model outputs, a task that is critical in applications requiring continuous feedback or reward-based optimization, such as reinforcement learning (RL) or other decision-making frameworks.

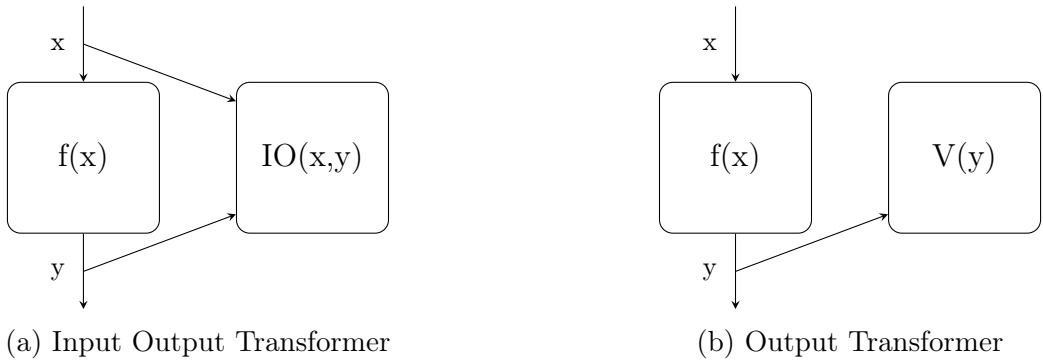
\begin{figure}
    \centering
    \begin{subfigure}[b]{0.48\textwidth}
        \centering
        \begin{tikzpicture}
            \node[draw, rectangle, minimum size=2cm, rounded corners=5pt] (square1) {f(x)};
            \draw[-stealth] ($(square1.north)+(0,1)$) -- ($(square1.north)+(0,0.0)$);
            \node[left, font=\fontsize{10}{12}\selectfont] at ($(square1.north)+(-.1,0.5)$) {x};
            \draw[-stealth] ($(square1.south)+(0,0)$) -- ($(square1.south)+(0,-1)$);
            \node[left, font=\fontsize{10}{12}\selectfont] at ($(square1.south)+(-.1,-0.5)$) {y};
            \node[draw, rectangle, minimum size=2cm, right=1cm of square1, rounded corners=5pt] (square2) {IO(x,y)};
            \draw[-stealth] ($(square1.north)+(0.0,0.5)$) -- ($(square2.west)+(0.0,0.75)$);
            \draw[-stealth] ($(square1.south)+(0,-0.5)$) -- ($(square2.west)+(0,-0.75)$);
        \end{tikzpicture}
        \caption{Input Output Transformer}
    \end{subfigure}
    \hfill
    \begin{subfigure}[b]{0.48\textwidth}
        \centering
        \begin{tikzpicture}
            \node[draw, rectangle, minimum size=2cm, rounded corners=5pt] (square1) {f(x)};
            \draw[-stealth] ($(square1.north)+(0,1)$) -- ($(square1.north)+(0,0.0)$);
            \node[left, font=\fontsize{10}{12}\selectfont] at ($(square1.north)+(-.1,0.5)$) {x};
            \draw[-stealth] ($(square1.south)+(0,0)$) -- ($(square1.south)+(0,-1)$);
            \node[left, font=\fontsize{10}{12}\selectfont] at ($(square1.south)+(-.1,-0.5)$) {y};
            \node[draw, rectangle, minimum size=2cm, right=1cm of square1, rounded corners=5pt] (square2) {V(y)};
            \draw[-stealth] ($(square1.south)+(0,-0.5)$) -- ($(square2.west)+(0,-0.75)$);
        \end{tikzpicture}
        \caption{Output Transformer}
    \end{subfigure}
    \caption{Comparison of the Presented Reward Models}
\end{figure}

In this work, we introduce two transformer-based architectures designed to serve as reward models: the Input-Output Transformer (IO Transformer) and the Output Transformer. These models are capable of assessing the quality of a model’s predictions by evaluating the relationship between the input and the output. Transformers have been extensively applied in supervised learning tasks. However, their potential to generate nuanced feedback as part of a reward function remains underexplored. Our reward models aim to fill this gap, providing more precise evaluations that account for both the quality of the input and its influence on the output.

Our experiments show that the IO Transformer and Output Transformer can accurately evaluate model performance across a range of vision tasks, with a particular focus on binary image segmentation. In this task, the quality of the output, such as a segmentation mask, is often highly dependent on the input, making it an ideal use case for input-output reward models. The IO Transformer, which assesses both the input and output, excels in scenarios where the output is sensitive to variations in the input. On the other hand, the Output Transformer focuses solely on evaluating the output, making it suitable for applications where the input variability is minimal or irrelevant.

This work contributes to the broader application of transformers in computer vision by introducing architectures designed to provide richer, more context-aware feedback. Our results demonstrate that the IO Transformer and Output Transformer can deliver state-of-the-art evaluation accuracy on tasks where precise feedback is critical. These reward models not only outperform traditional value networks in input-dependent tasks, but also offer the potential for future integration with reinforcement learning methods to optimize policies in complex environments, such as segmentation tasks.

By presenting the IO Transformer and Output Transformer, we aim to expand the use of vision transformer architectures beyond object detection and segmentation tasks, opening new avenues for their application in reward-driven optimization and quality assessment. This work serves as a foundation for future research into the integration of transformer-based reward models with reinforcement learning frameworks, where more nuanced, context-sensitive feedback is required to improve decision-making and policy learning.

\section{Relevant Work}

\subsection{Transformers in Computer Vision}
Transformers, initially introduced for natural language processing (NLP) tasks by Vaswani et al. \cite{vaswani2017attention}, have significantly impacted computer vision. The Vision Transformer (ViT) \cite{dosovitskiy2020image} demonstrated that treating images as sequences of patches could match or outperform traditional convolutional neural networks (CNNs) on standard vision benchmarks. However, ViTs posed challenges, such as the computationally expensive quadratic complexity $\mathcal{O}(n^2 \cdot d)$—where $n$ is the total number of patches and $d$ is the feature dimension—for each self-attention layer, as well as poor performance \cite{touvron2021training} on datasets smaller than the extremely large ImageNet-21k \cite{deng2009imagenet} or JFT-300M \cite{dosovitskiy2020image}.

To address these limitations, Liu et al.  introduced the Swin Transformer \cite{liu2021swin}, a hierarchical model with local window-based attention, achieving improved scalability by reducing self-attention complexity to a linear complexity $\mathcal{O}(n \cdot w^2 \cdot d)$, where $w$ is the window size. This architecture has become widely adopted for tasks requiring fine-grained predictions, such as image segmentation \cite{hatamizadeh2021swin} and image restoration \cite{liang2021swinir}. Later, SwinV2 \cite{liu2022swin} improved on these models by introducing post-normalization techniques and positional bias mechanisms, which ensured stable training on high-resolution datasets. The innovations provided by the Swin and SwinV2 laid the foundation for efficient and scalable models in both supervised \cite{conde2022swin2sr} and reinforcement learning-based \cite{meng2024deep} vision tasks.

\subsection{Reinforcement Learning in Computer Vision Tasks}
RL has been applied to various computer vision tasks, such as object detection, robotic control, and visual tracking. The introduction of Deep Q-Networks (DQN) \cite{mnih2013playing} and Asynchronous Advantage Actor-Critic (A3C) \cite{mnih2016asynchronous} showcased the potential of RL for training agents that interact with visual environments. Pathak et al. \cite{pathak2017curiosity} further advanced the field by incorporating curiosity-driven exploration, which helps agents learn efficiently in environments with sparse rewards. More recently, model-based RL approaches have demonstrated improved generalization and sample efficiency by planning with learned models \cite{hubert2021learning}.

Despite these developments, RL architectures often rely on CNN-based methods for feature extraction, which limits their ability to capture fine-grained, pixel-level information critical for tasks such as segmentation and object tracking \cite{mnih2015human, zhang2017deep}. This highlights the need for the development of hybrid approaches that combine RL with advanced feature extractors, such as transformers, to better address high-dimensional visual tasks \cite{parisotto2020stabilizing, chen2021decision}.

Although our work draws conceptual inspiration from reinforcement learning, particularly the actor-critic method \cite{sutton2018reinforcement}, we diverge from traditional implementations that rely on time-step-based interactions and value function approximations. The actor-critic method typically employs two networks: an "actor" that makes decisions and a "critic" that evaluates those decisions to guide future actions. Our architecture, while inspired by this framework, is not directly integrated into a reinforcement learning setup. Instead, we focus on leveraging transformers as reward models to directly assess the quality of outputs without the need for intermediate value approximations or temporal feedback.

\subsection{Siamese Networks and Transformer Hybrids}
Siamese networks were first introduced by Bromley et al. \cite{bromley1993signature} for signature verification. These networks consist of twin models with shared weights, designed to compare two inputs by learning their similarity. They have since been applied to diverse tasks, including face recognition, change detection, and medical imaging \cite{koch2015siamese, zagoruyko2015learning}. The architecture’s ability to map similar inputs to proximate locations in feature space makes it highly effective for comparison-based tasks \cite{wu2018unsupervised}.

Recent work has explored the integration of transformers into Siamese networks. Bandara et al. \cite{bandara2022transformer} introduced a transformer-based Siamese network for change detection, achieving state-of-the-art performance on the LEVIR-CD and DSIFN-CD datasets. Yu et al.  proposed TransMatch \cite{yu2020transmatch}, a hybrid architecture combining Siamese networks with transformer encoders for cross-modal matching. Our IO Transformer architecture draws inspiration from these models, employing separate SwinV2 encoders for input and output images. Unlike traditional Siamese networks, we decouple the weights between the two encoders, allowing the model to capture more nuanced input-output relationships through cross-attention layers.

\subsection{Cross-Attention for Input-Output Evaluation}
Cross-attention mechanisms have been instrumental in tasks requiring the fusion of multiple inputs, such as visual question answering (VQA) \cite{lu2019vilbert} and multimodal learning \cite{alayrac2022flamingo}. The ability of cross-attention to selectively focus on relevant parts of both inputs makes it particularly suitable for evaluating complex relationships. In the context of our work, cross-attention connects the input and output encoders, enabling the model to provide more precise evaluations of output quality in relation to the input.

Our approach is inspired by research in multimodal transformers, such as Flamingo \cite{alayrac2022flamingo}, which use cross-attention to integrate information from different modalities. However, while Flamingo focuses on fusing image and text inputs, our IO Transformer applies cross-attention solely to input-output pairs. This allows the reward model to generate nuanced evaluations tailored to the specific needs of vision tasks, such as image segmentation. The use of cosine similarity in our cross-attention layers aligns with that of SwinV2, ensuring that the dimensionality remains consistent throughout the model.

\subsection{Summary and Identification of Research Gaps}
The rise of transformers and reward models has significantly expanded the capabilities of computer vision and RL architectures. However, there is still a need for research on architectures that can evaluate input-output dependencies effectively. While previous work has explored the use of reward models in language generation and robotic control, there is limited research on their application to vision tasks \cite{christiano2017deep, ouyang2022training}. Furthermore, most existing RL models rely on CNN-based feature extraction, which limits their applicability to tasks requiring detailed visual understanding.

Our work addresses these gaps by proposing the IO Transformer, a novel reward model architecture that leverages SwinV2-based encoders to evaluate both input and output quality. This approach offers more reliable feedback for policy optimization and paves the way for more adaptive and reliable RL systems in computer vision.

\section{Method}

\begin{figure}[htbp]

\centering
\subfloat[IO V8 \\ Transformer]{
  \begin{tikzpicture}[
    scale=0.5,
    every node/.style={scale=0.5},
    encoder/.style={rectangle, draw, fill=white, text width=8cm, text centered, minimum height=3cm, rounded corners=3pt},
    cross/.style={rectangle, draw, fill=white, text width=2.5cm, text centered, minimum height=2.5cm, rounded corners=3pt},
    line/.style={draw, ->, >=latex}
  ]
    \node[encoder] (upper) at (0,2.5) {\makebox[7.5cm][c]{Swinv2 Input Encoder}\hfill S5};
    \node[encoder] (lower) at (0,-2.5) {\makebox[7.5cm][c]{Swinv2 Output Encoder}\hfill S5};
    \node[cross, right=1.25cm of upper.east, yshift=-2.5cm] (cross) {Cross Attention};
    \draw[line] (upper.east) -- ++(2.525cm,-1.875);
    \draw[line] (lower.east) -- ++(2.525cm,1.875);
  \end{tikzpicture}
}
\hfill
\subfloat[IO W12 \\ Transformer]{
  \begin{tikzpicture}[
    scale=0.5,
    every node/.style={scale=0.5},
    block/.style={rectangle, draw, fill=white, text width=1cm, text centered, minimum height=0.5cm, rounded corners=2pt},
    encoder/.style={rectangle, draw, fill=white, text width=8cm, text centered, minimum height=3cm, rounded corners=4pt},
    decoder/.style={rectangle, draw, fill=white, text width=1.5cm, text centered, minimum height=1.5cm, rounded corners=2pt},
    line/.style={draw, ->, >=latex}
  ]
    \node[encoder] (upper) at (0,2.5) {Swinv2 Input Encoder};
    \node[encoder] (lower) at (0,-2.5) {Swinv2 Output Encoder};
    \node[block] at ($(upper.south)!0.5!(lower.north) + (-3.5,0)$) (cross1) {cross};
    \node[block, right=0.2cm of cross1] (patch1) {patch emb};
    \node[block, right=0.2cm of patch1] (cross2) {cross x2};
    \node[block, right=0.2cm of cross2] (patch2) {patch emb};
    \node[block, right=0.2cm of patch2] (cross3) {cross x2};
    \node[block, right=0.2cm of cross3] (patchmerge) {patch merge};
    \node[block, right=0.2cm of patchmerge] (cross4) {cross x2};
    \node[block, right=0.2cm of cross4] (cross5) {cross x2};
    \path [line] (cross1) -- (patch1);
    \path [line] (patch1) -- (cross2);
    \path [line] (cross2) -- (patch2);
    \path [line] (patch2) -- (cross3);
    \path [line] (cross3) -- (patchmerge);
    \path [line] (patchmerge) -- (cross4);
    \path [line] (cross4) -- (cross5);
    \foreach \x/\block in {1/cross1,2/cross2,3/cross3,4/cross4,5/cross5} {
      \path [line] ($(upper.south west)!{0.125+0.1875*(\x-1)}!(upper.south east)$) -- (\block.north);
      \path [line] ($(lower.north west)!{0.125+0.1875*(\x-1)}!(lower.north east)$) -- (\block.south);
      \node[anchor=south] at ($(upper.south west)!{0.125+0.1875*(\x-1)}!(upper.south east)$) {S\x};
      \node[anchor=north] at ($(lower.north west)!{0.125+0.1875*(\x-1)}!(lower.north east)$) {S\x};
    }
    \node[decoder, right=0.5cm of cross5] (mlp) {MLP};
    \path [line] (cross5) -- (mlp.west);
  \end{tikzpicture}
}
\caption{}
\label{fig:model_architectures}
\end{figure}

\subsection{IO Transformer}

To effectively evaluate the input-output dependency in various computer vision tasks, we propose the IO Transformer. This architecture leverages two independent SwinV2-based encoders to process both the input and output data streams separately. Traditional Siamese networks use shared encoders, but our approach deliberately decouples the weights between the two SwinV2 encoders. This separation allows the model to capture finer details specific to each data stream, enhancing the ability to evaluate output quality relative to the input \cite{bromley1993signature, bandara2022transformer}.

The input encoder specializes in processing raw inputs (e.g., original images), while the output encoder focuses on evaluating the outputs generated by the actor model (e.g., segmentation masks). The combination of these dual encoders through cross-attention layers ensures a nuanced evaluation of how well the output aligns with the input conditions.

\subsubsection{Cross-Attention Mechanism}

To integrate the features generated by the input and output encoders, we use cosine-based cross-attention layers, which preserve the attention consistency from the SwinV2 backbones \cite{liu2022swin}. In these cross-attention layers, the query vector \( Q \) comes from the output encoder, while the key and value vectors \( K \) and \( V \) are derived from the input encoder. This design ensures that the output features attend to relevant aspects of the input, allowing for precise evaluation. 

The cross-attention operation follows:

\begin{equation}
\text{CrossAttention}(I_i, O_i) = \text{softmax}\left(\frac{\cos(Q(O_i), K(I_i))}{\tau} + B\right) V(I_i)
\end{equation}

Where:
\begin{itemize}
    \renewcommand{\labelitemi}{--}
    \item \( I_i \) and \( O_i \) are features from the input and output encoders, respectively.
    \item \( \tau \) is a learned temperature parameter controlling the distribution sharpness.
    \item \( B \) is the relative positional bias, ensuring the model retains spatial consistency.
\end{itemize}

This formulation ensures that the reward model captures the relationship between the input and output, essential in tasks where output quality heavily depends on input conditions \cite{zhu2020deformable}. For example, in image segmentation, the clarity of the input (e.g., lighting or noise) significantly impacts the segmentation mask quality, which the IO Transformer accurately evaluates through its input-output attention mechanism.

\subsubsection{Comparison to Siamese Networks}

Our architecture diverges from traditional Siamese networks by using two fully independent SwinV2 encoders. This design allows each encoder to specialize in its respective role, avoiding the limitations imposed by weight sharing in Siamese architectures. In comparison, shared weights can constrain the model’s ability to adapt to differences between inputs and outputs, particularly in tasks with high variability, such as binary segmentation \cite{koch2015siamese}. By decoupling the encoders, the IO Transformer delivers better performance in scenarios where the relationship between input and output is non-trivial.

\subsection{Output Transformer: Design and Implementation}

The Output Transformer architecture is built using a SwinV2 backbone, focusing exclusively on evaluating the model's outputs. In contrast to the IO Transformer, which leverages input-output relationships, the Output Transformer operates on the assumption that the output alone provides sufficient information for accurate evaluation. This design makes it ideal for applications where the variability in input has minimal effect on the quality of the output, such as predictive quality checks or isolated feature evaluations.

\subsubsection{Architectural Variants}

We developed two versions of the Output Transformer:
\begin{itemize}
    \renewcommand{\labelitemi}{--}
    \item \textbf{SwinV2 Output Transformer:} This version relies on the original SwinV2 backbone, trained with minimal architectural changes. It is lightweight and effective when fine-tuned on simpler binary classification tasks.
    \item \textbf{Custom-Layer SwinV2 Output Transformer:} This enhanced version introduces additional layers at the end of the SwinV2 backbone, including self-attention layers and MLP layers to extract finer features from the output data.
\end{itemize}

Each version of the Output Transformer follows a single-stream processing pipeline:
\begin{enumerate}
    \item The model receives the output from the actor model as input.
    \item This output is encoded using the SwinV2 backbone into feature embeddings.
    \item Additional layers (if included) further refine these embeddings before they are passed to the final classification or evaluation head.
\end{enumerate}

\subsubsection{Advantages over Input-Output Architectures}

In scenarios where input variability is minimal or irrelevant, the Output Transformer offers several key advantages:

\begin{itemize}
    \renewcommand{\labelitemi}{--}
    \item \textbf{Lower computational overhead:}  Since it processes only the outputs, the Output Transformer requires fewer computational resources than input-output models.
    \item \textbf{Simplicity:} Training and deployment are streamlined, as there is no need to align input-output pairs.
    \item \textbf{Robust evaluation:}  The model excels in tasks like mask refinement, feature verification, or object categorization, where the output carries all necessary information.
\end{itemize}

\subsubsection{Limitations and Challenges}

While the Output Transformer is effective in scenarios with stable inputs, it presents certain limitations:
\begin{itemize}
    \renewcommand{\labelitemi}{--}
    \item \textbf{Inability to handle input-output dependencies:}   It fails when the output quality is tightly linked to input conditions (e.g., underexposed images in segmentation tasks).
    \item \textbf{Over-reliance on actor performance:} The model’s reward signal is directly dependent on the quality of the actor model's outputs, making it vulnerable to noisy predictions.
\end{itemize}

\subsubsection{Future Directions for Output-Only Models}

Future work could explore the integration of RLHF (Reinforcement Learning from Human Feedback) with the Output Transformer. By aligning automated reward signals with human preferences, the model could become more versatile and reliable, particularly in sensitive applications like medical diagnostics or autonomous driving.

Additionally, hybrid architectures that switch between Output-Only and IO-based evaluation modes could be developed, allowing models to dynamically adapt their reward mechanisms based on task requirements.

\section{Experiments}

\subsection{Model Task}
To thoroughly evaluate the potential of our proposed architectures as critic networks, we focus on binary image classification and dual-image binary classification tasks. Binary classification is an ideal choice for several reasons:

\begin{enumerate}
    \item \textbf{Simplicity:} The task's inherent simplicity allows us to focus on the architectures' performance without the added complexity of multi-class problems. 
    \item \textbf{Versatility:} Binar classification has broad applications, including content realism detection, medical image pre-screening, online content moderation, and product quality control \cite{haarnoja2018soft}, \cite{ouyang2022training}.
    \item \textbf{Data synthesizability:} Generating synthetic datasets for binary classification is straightforward, allowing for large-scale evaluation \cite{mnih2016asynchronous}. This ensures consistency during data preparation and aids in model training.
\end{enumerate}

\subsection{Training Data}

We generate a custom dataset called the IO Segmentation Dataset by leveraging segmentation masks produced from approximately 40,000 images. These images are processed with an image segmentation model, which provides high-quality masks of image foregrounds. Since segmentation models often produce imprecise masks, each output undergoes a hand classification step to remove improper masks from the training data.

To capture input-output relationships, the input and output images are concatenated during evaluation, following methods employed in dual-input models such as Siamese networks \cite{koch2015siamese}.

Our models use SwinV2 Large and SwinV2 Base as the primary backbones, pre-trained on ImageNet-22k and fine-tuned on ImageNet-1k \cite{liu2022swin}.

\subsection{Pre-training Data}
Given the complexity of our architecture, we incorporate a pre-training stage to align each model's components before fine-tuning on task-specific data. However, expanding the IO Segmentation Dataset to the required 100,000 images proved impractical. To address this, we developed the Change Dataset 25 (CD25), containing 100,000 images across 25 categories and five data types:
\begin{itemize}
    \renewcommand{\labelitemi}{--}
    \item \textbf{Anime}: Stylized character images commonly used in visual media.      
    \item \textbf{Cartoon}: Simplified, exaggerated depictions often appearing in animations.  
    \item \textbf{Food}: Dishes and ingredients in various environments.
    \item \textbf{Real-world places}: Images of cities, landscapes, and natural environments.  
    \item \textbf{Human faces}: Portraits and facial expressions in diverse contexts.
\end{itemize}

By creating permutations of these categories (e.g., cartoon as input and food as output), the pre-training step allows the models to learn input-output relationships \cite{bandara2022transformer}. This aligns with techniques used in visual comparison models like TransMatch \cite{yu2020transmatch}.

\subsection{Implementation}
All experiments were implemented in PyTorch and run on a system equipped with 6 Nvidia RTX 3090 GPUs. We used Microsoft’s Swin Transformer training loop, with several modifications for data augmentation and optimization strategies:

\begin{itemize}
    \renewcommand{\labelitemi}{--}
    \item Mixup augmentation \cite{zhang2017deep} was applied to the input data to reduce overfitting.    
    \item AdamW optimizer was used due to its superior convergence properties when compared to Adam \cite{loshchilov2017decoupled}.  
    \item  We employed a cosine learning rate scheduler during warm-up before switching to a linear scheduler.
\end{itemize}

Training hyperparameters:
\begin{itemize}
    \renewcommand{\labelitemi}{--}
    \item Image size: 256×256
    \item Learning rate: 2e-05 (base), 2e-07 (minimum), 2e-08 (warm-up)
\end{itemize}

\subsection{Results with IO Segemention Dataset}
\begin{figure}[H]
  \centering
  \includegraphics[width=0.8\textwidth]{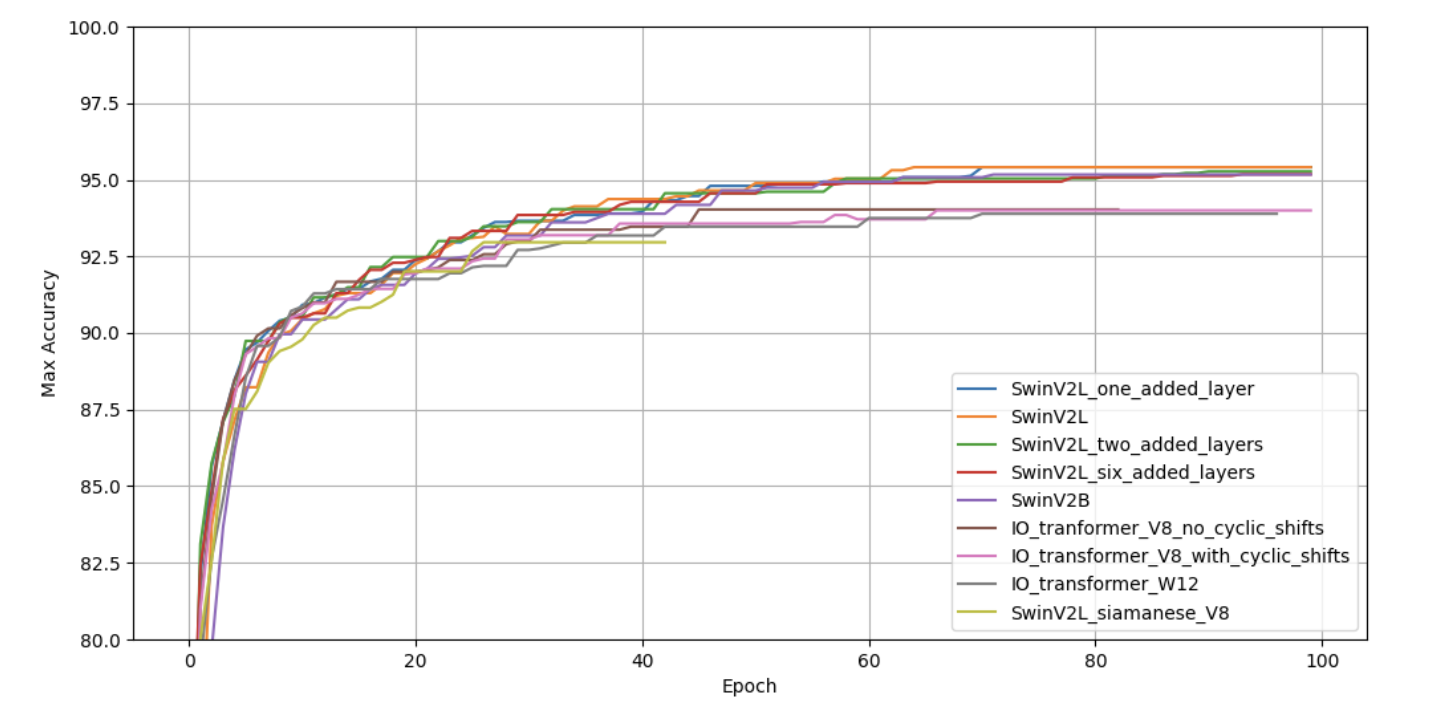}
  \caption{Max accuracy over epochs trained}
  \label{fig:your-label}
\end{figure}

\begin{table}[H]
  \centering
  \small
  \caption{Model Performance Comparison  (Sorted by Accuracy)}
  \label{tab:model_comparison_cyclic_sorted}
  \begin{tabular}{|r|c|c|c|l|}
    \hline
    \multicolumn{1}{|c|}{\textbf{Params}} & \textbf{Epochs} & \textbf{Accuracy} & \textbf{Cyclic Shift} & \multicolumn{1}{c|}{\textbf{Model}} \\
    \multicolumn{1}{|c|}{(in million)} & \textbf{Trained} & \textbf{(\%)} & \textbf{In Cross} & \\
    \hline
    224 & 100/100 & 95.41 & N/A & SwinV2 1 layer add \\
    195 & 100/100 & 95.41 & N/A & SwinV2 Large\\
    224 & 100/100 & 95.41 & N/A & SwinV2 1 layer add \\
    252 & 100/100 & 95.27 & N/A & SwinV2 2 layers add \\
    365 & 100/100 & 95.18 & N/A & SwinV2 6 layers add \\
    87 & 100/100 & 95.17 & N/A & SwinV2 Base\\
    250 & 83/100 & 94.03 & No & IO V8 Transformer \\
    250 & 100/100 & 94.00 & Yes & IO V8 Transformer \\
    296 & 98/100 & 93.89 & No & IO W12 Transformer \\
    195 & 30/30 & 93.71 & N/A & SwinV2 Large (Microsoft fine tuning loop)\\
    250 & 63/100 & 93.38 & Yes & IO V8 Transformer pre-trained on CD25 \\
    87 & 30/30 & 93.30 & N/A & SwinV2 Base (Microsoft fine tuning loop) \\
    365 & 43/100 & 92.96 & No & Siamese IO Transformer  \\
    \hline
  \end{tabular}
\end{table}

The IO Transformer models performed competitively across different architectures. Notably, the Output Transformer achieved an accuracy of \SI{95.41}{\percent} using SwinV2-based backbones, as shown in Table 1 above. Our experiments confirm findings from SimMIM \cite{xie2022simmim} that adding attention layers to the model head does not always improve performance.
For Swin models parameters the larger the number parameters added to the end the worse accuracy observed. For Swin models, adding more parameters to the end of the model results in worse accuracy, and the more parameters added, the worse the accuracy gets.

Key observations:

\begin{itemize}
    \renewcommand{\labelitemi}{--}
    \item Cyclic shifts in cross-attention layers slightly degraded the model’s performance.
    \item Pre-training on CD25 led to a slight decrease in performance, suggesting that domain-specific fine-tuning is essential for the IO Transformer’s success.
\end{itemize}

\subsection{Results with CD25} 

We evaluated two IO Transformer variants on the CD25 dataset. The IO V8 Transformer achieved \SI{100}{\percent} accuracy, demonstrating its effectiveness in scenarios where the output is entirely dependent on the input.

\begin{table}[H]
  \centering
  \small
  \caption{Model Performance Comparison  (Sorted by Accuracy)}
  \label{tab:model_comparison_io}
  \begin{tabular}{|r|c|c|c|l|}
    \hline
    \multicolumn{1}{|c|}{\textbf{Params}} & \textbf{Epochs} & \textbf{Accuracy} & \textbf{Cyclic Shift} & \multicolumn{1}{c|}{\textbf{Model}} \\
    \multicolumn{1}{|c|}{(in million)} & \textbf{Trained} & \textbf{(\%)} & \textbf{In Cross} & \\
    \hline
    250 & 69/100 & 100 & no & IO V8 Transformer\\
    325 & 23/100 & 99.2 & yes & IO W12 Transformer\\
    \hline
  \end{tabular}
\end{table}

The results indicate that the IO Transformer excels when the task requires precise input-output dependency analysis. In such cases, pre-trained models benefit significantly from attention-based architectures like SwinV2. Further research could explore hybrid architectures to bridge the performance gap between IO and Output Transformers.

\section{Conclusion}

In this paper, we presented the IO Transformer and Output Transformer architectures, designed to address challenges in reward modeling within computer vision. Our results demonstrate that the IO Transformer excels when input-output dependencies are critical, achieving perfect accuracy on the CD25 dataset. Conversely, the Output Transformer is more effective in scenarios where input variability is minimal, as reflected by its \SI{95.41}{\percent} accuracy on the IO Segmentation Dataset.

The IO Transformer offers nuanced evaluations by coupling input-output analysis, making it well-suited for high-stakes applications such as medical imaging and autonomous vehicles. However, its computational complexity highlights the trade-off between detailed reward modeling and resource efficiency. On the other hand, the Output Transformer provides a lightweight alternative for use cases with stable input conditions, showcasing its potential for deployment in real-time inference systems.

Our work highlights the importance of aligning reward model selection with task-specific requirements. Future research should focus on bridging the gap between the two architectures through hybrid models, capable of dynamically adapting to changing input-output dependencies. Furthermore, incorporating Reinforcement Learning from Human Feedback (RLHF) may enhance the adaptability and reliability of these models in complex environments.

In conclusion, this paper advances the understanding of reward-driven policy updates in computer vision, laying the groundwork for more adaptive, stable, and efficient reinforcement learning frameworks. We hope this work inspires further exploration of architectures that can approximate reward functions more accurately across diverse domains.

\newpage
\section{References}
\bibliographystyle{unsrt}
\bibliography{references}

\end{document}